\documentclass[sigconf]{acmart}
\AtBeginDocument{%
  }

\usepackage{latexsym}
\usepackage{graphicx}
\usepackage{tcolorbox}
\usepackage{booktabs}
\usepackage{multirow}
\usepackage{makecell}
\setcopyright{rightsretained}
\copyrightyear{2025}
\acmYear{2025}
\acmConference[Genaiecom '25]{Proceedings of the second workshop on Generative AI for E-Commerce}{September 22, 2025}{Prague, CZ}
\acmBooktitle{Proceedings of the second workshop on Generative AI for E-Commerce 2025, September 22, 2025}
\acmYear{2025}
\acmMonth{09}
\acmDOI{}
\acmISBN{}
\acmPrice{15.00}

\begin{document}

\title{CRMAgent: A Multi-Agent LLM System for E-Commerce \\ CRM Message Template Generation}

% \affiliation{%

\author{Yinzhu Quan}
\affiliation{%
  \institution{Georgia Institute of Technology}
  \city{Atlanta}
  \state{Georgia}
  \country{USA}}
\email{yquan9@gatech.edu}

\author{Xinrui Li}
\affiliation{%
  \institution{ByteDance Inc.}
  \city{Seattle}
  \state{Washington}
  \country{USA}}
\email{ava.li@bytedance.com}

\author{Ying Chen}
\affiliation{%
  \institution{ByteDance Inc.}
  \city{San Jose}
  \state{California}
  \country{USA}}
\email{eric.ychen@bytedance.com}

\renewcommand{\shortauthors}{Quan et al.}

\begin{abstract}
In e-commerce private-domain channels such as instant messaging and e-mail, merchants engage customers directly as part of their Customer Relationship Management (CRM) programmes to drive retention and conversion. While a few top performers excel at crafting outbound messages, most merchants struggle to write persuasive copy because they lack both expertise and scalable tools. We introduce CRMAgent, a multi-agent system built on large language models (LLMs) that generates high-quality message templates and actionable writing guidance through three complementary modes. First, group-based learning enables the agent to learn from a merchant's own top-performing messages within the same audience segment and rewrite low-performing ones. Second, retrieval-and-adaptation fetches templates that share the same audience segment and exhibit high similarity in voucher type and product category, learns their successful patterns, and adapts them to the current campaign. Third, a rule-based fallback provides a lightweight zero-shot rewrite when no suitable references are available. Extensive experiments show that CRMAgent consistently outperforms merchants' original templates, delivering significant gains in both audience-match and marketing-effectiveness metrics.
\end{abstract}





\maketitle

%%%%%%%%%%%%%%%%%%%%%%%%%%%%%%
\section{Introduction}

Customer Relationship Management (CRM) \cite{kumar2018customer} messages serve as a primary lever for user retention and revenue generation on modern e-commerce platforms. Merchants routinely push promotional notifications that announce new products, highlight price drops, or deliver loyalty rewards to segmented audiences through app and messaging channels. Crafting effective templates remains a bottleneck because most merchants lack dedicated marketing expertise, work under tight time constraints, and therefore rely on generic system-generated copy. Many CRM messages consequently underperform, leaving substantial engagement and conversion potential untapped.

\begin{figure*}[h!]
    \centering
    \includegraphics[width=0.8\linewidth]{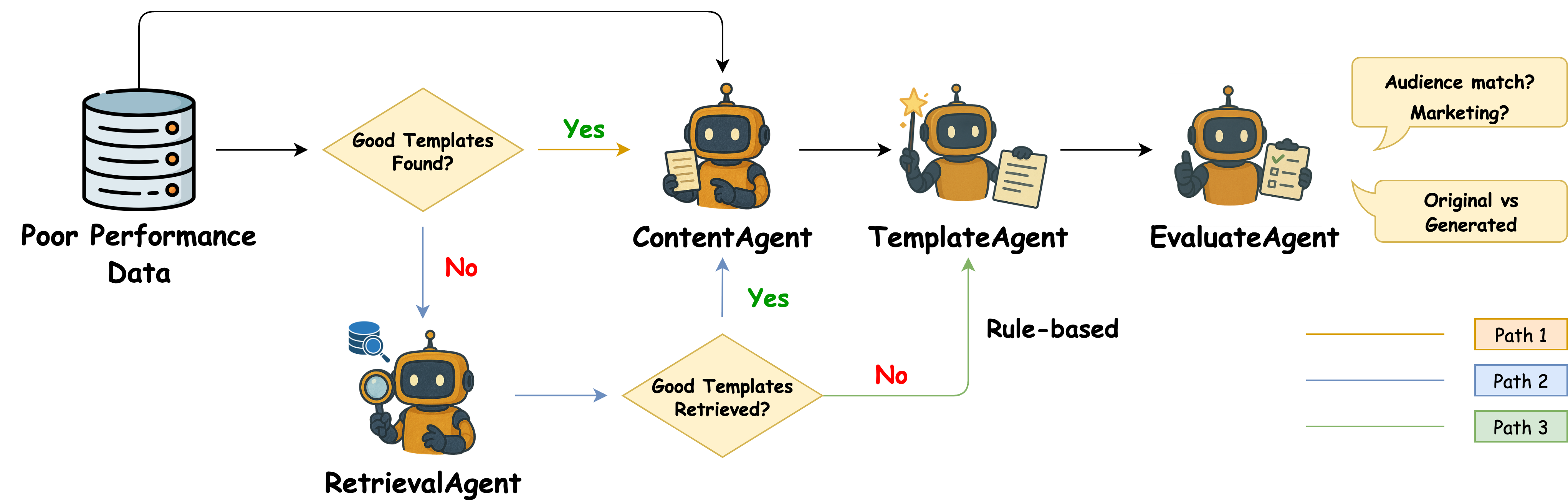}
    \caption{Framework of CRMAgent, a multi-agent system for improving underperforming CRM message templates in e-commerce. The system begins with a set of low-performing message templates and checks whether any strong templates exist for the same merchant and audience segment. In path 1, when in-group top-performing exemplars are available, \texttt{ContentAgent} analyzes them to extract key persuasive elements that guide \texttt{TemplateAgent} in rewriting the original message. In path 2, if no internal top-performing exemplars exist, \texttt{RetrievalAgent} searches cross-merchant templates that share the same audience profile and exhibit high similarity in voucher type and product category. Valid retrieved templates are passed through \texttt{ContentAgent} for insight extraction and then rewritten. In path 3, if no suitable templates are found, the system falls back to a rule-based rewriting strategy. In all cases, \texttt{EvaluateAgent} supports two complementary evaluation strategies: fine-grained scoring of each message and pairwise preference judgment between the original and the rewritten version.}
    \label{fig:framework}
\end{figure*}

Recent work has shown the effectiveness of LLM agents for structured decision-making in operational domains such as logistics~\cite{ieva2025enhancing,quan2025leveraging}, supply chain management~\cite{li2023large,quan2024invagent,wang2025large}, and economic reasoning~\cite{quan2024econlogicqa,guo2024econnli,liu2025econwebarena}. To address this performance gap in CRM, we present CRMAgent (Figure~\ref{fig:framework}), a multi-agent system built on large language models (LLMs) that automatically improves underperforming CRM message templates in e-commerce. Using historical engagement data, CRMAgent identifies low-performing messages, retrieves high-quality examples from similar contexts, and produces improved variants tailored to each merchant and audience segment group. The workflow is divided among four specialized agents: \texttt{ContentAgent}, \texttt{RetrievalAgent}, \texttt{TemplateAgent}, and \texttt{EvaluateAgent}, each responsible for a distinct subtask.

Our contributions are threefold: (1) We present an end-to-end multi-agent LLM framework for e-commerce CRM message template generation. The system decomposes the task into four specialized agents: \texttt{ContentAgent}, \texttt{RetrievalAgent}, \texttt{TemplateAgent} and \texttt{EvaluateAgent}, which together handle data ingestion, analysis, template generation and automatic quality assessment. (2) To cover diverse data scenarios, we develop a \texttt{TemplateAgent} which can learn from a merchant's own high-performing messages within the same audience segment (group-based learning), retrieve and adapt cross-merchant exemplars that match audience segment, voucher type and product category (retrieval-and-adaptation), and fall back to a rule-based zero-shot rewrite when no suitable references exist. (3) And we implement \texttt{EvaluateAgent} which supports two evaluation modes: assigning separate scores for audience fit and marketing effectiveness to each candidate, and directly selecting the superior message.

%%%%%%%%%%%%%%%%%%%%%%%%%%%%%%
\section{Related Work}

Recent multi-agent LLM systems adopt modular architectures to address structured generation tasks across diverse domains. For instance, \citet{singh2024personal} design role-based agents for collaborative travel planning, while DocAgent~\cite{yang2025docagent} structures code documentation into sequential reader–writer–verifier stages. MAxPrototyper~\cite{yuan2024maxprototyper} coordinates theme, text, and visual elements under a central controller for UI prototyping, and InvAgent~\cite{quan2024invagent} demonstrates that agent specialization benefits inventory management in logistics. Similarly, CRMAgent applies a task-specialized multi-agent framework to CRM message rewriting, with agents dedicated to content analysis, retrieval, generation, and evaluation. Complementary to modularity, retrieval-augmented generation (RAG) improves LLM output by grounding in exemplar data or structured knowledge. Retrieve-and-Edit~\cite{hashimoto2018retrieve}, PEARL~\cite{malik2024pearl}, and \citet{chen2024knowledge} leverage retrieval to enhance preference modeling and factual accuracy. Recent work also explores temporal event modeling and retrieval~\cite{liu2024tpp,liu2025retrieval}, combining LLMs with temporal point processes to unify event semantics and timing, while \citet{liu2024fine} show that fine-tuning with retrieved exemplars improves personalization. Extending these directions, CRMAgent retrieves high-performing CRM templates based on audience, product, and voucher metadata, and generates tailored messages using marketing feedback.

%%%%%%%%%%%%%%%%%%%%%%%%%%%%%%
\section{Multi-Agent Collaboration for CRM Message Template Generation}

This section presents the architecture of the CRMAgent system, detailing the roles and interactions of its four specialized agents.

%%%%%%%%%%%%%%%%%%%%%%%%%%%%%%
\subsection{\texttt{ContentAgent}}

\texttt{ContentAgent} performs structured diagnosis of CRM message templates by comparing high- and low-performing examples within the same audience segment. It summarizes why certain messages achieved better engagement than others, focusing on structural and stylistic patterns in content design. Specifically, it analyzes differences in phrasing, organization, targeting logic, and the way incentives are presented. To ensure contextual relevance, engagement percentiles are calculated within each audience segment, defined by the plan's audience label. Figure~\ref{fig:prompt_content_agent} shows the prompt used to guide this comparative analysis.

\begin{figure}[h!]
  \centering
  \begin{tcolorbox}[title=\textbf{Prompt for \texttt{ContentAgent}}, colback=gray!5, colframe=gray!40!black, fontupper=\small]

    You are a CRM marketing strategist. Below are several CRM message templates sent to the audience segment \texttt{\{audience\}}. Each message promotes different categories and offers, with either good or poor engagement. Your task is to identify clear, actionable reasons why some templates performed better than others. Do not focus on personalization customization. Analyze and summarize the differences between high- and low-performing messages. Be concise and analytical. Focus on patterns in tone, structure, targeting, and incentive design.\\

    === High-Performing Templates ===\\
    \texttt{\{Template | Category | Voucher\}}\\
    
    === Low-Performing Templates ===\\
    \texttt{\{Template | Category | Voucher\}}

  \end{tcolorbox}
  \caption{Prompt used by \texttt{ContentAgent} to generate comparative analyses of CRM template performance.}
  \label{fig:prompt_content_agent}
\end{figure}

%%%%%%%%%%%%%%%%%%%%%%%%%%%%%%
\subsection{\texttt{RetrievalAgent}}

\texttt{RetrievalAgent} retrieves high-quality CRM message templates from other merchants within the same audience group, selecting those that are most similar in product category and voucher type. When no suitable high-performing templates exist within the same merchant-audience group, \texttt{RetrievalAgent} searches cross-merchant examples to support rewriting. Each candidate message is represented by a dense embedding generated using a SentenceTransformer \citep{reimers2019sentence} model all-MiniLM-L6-v2~\cite{wang2020minilm}. Embeddings are computed based on structured metadata, including the product category hierarchy and voucher attributes. These vectors are indexed using Faiss~\cite{douze2024faiss} for efficient similarity search. At inference time, the agent encodes the target message's metadata into a dense vector and retrieves the top-$k$ most similar templates to serve as few-shot references for \texttt{TemplateAgent}. If both product and voucher metadata are missing, the system skips \texttt{RetrievalAgent} and falls back to rule-based rewriting.

%%%%%%%%%%%%%%%%%%%%%%%%%%%%%%
\subsection{\texttt{TemplateAgent}}

\texttt{TemplateAgent} rewrites low-performing CRM message templates by combining diagnostic insights from \texttt{ContentAgent} with high-performing exemplars. These exemplars are either sourced from the same merchant and audience segment or retrieved across merchants by \texttt{RetrievalAgent}. The agent receives four types of input: the original message, a structured explanation of its weaknesses, a summary of success patterns, and one or more reference templates. To improve clarity, persuasiveness, and alignment with audience expectations while still preserving the original promotional intent such as product and voucher offers, \texttt{TemplateAgent} follows a three-stage template generation strategy: it first uses in-group exemplars if available, otherwise retrieves top-ranked examples from the same audience group with similar product and voucher characteristics, and finally generates a message using rule-based rewriting when no suitable references are found. The exemplar-guided prompt is shown in Figure~\ref{fig:prompt_template_agent}, and the rule-based fallback prompt is shown in Figure~\ref{fig:prompt_template_agent_rule_based}.

\begin{figure}[h!]
  \centering
  \begin{tcolorbox}[title=\textbf{Prompt for \texttt{TemplateAgent}}, colback=gray!5, colframe=gray!40!black, fontupper=\small]

    You are a CRM marketing copywriter. A past CRM message performed poorly. Below is the Original Poor Template:\\
    \texttt{\{Poor Performance Template\}}\\

    Problem Analysis:\\
    \texttt{\{Poor Performance Reason\}}\\

    Here are some high-performing templates from the same or similar campaigns:\\
    \texttt{\{Good Performance Template\}}\\

    Summary of what makes high-performing templates successful:\\
    \texttt{\{Good Performance Reason\}}\\

    Now, please rewrite the original message into a short, action-driven, persuasive push notification. Do not focus on personalization customization. Focus on improving clarity, structure, and incentive framing --- while preserving the original intent. Respond in the similar language style as the Original Poor Template.
  \end{tcolorbox}
  \caption{Prompt used by \texttt{TemplateAgent} to rewrite underperforming CRM messages using performance diagnostics and exemplar templates.}
  \label{fig:prompt_template_agent}
\end{figure}

\begin{figure}[h!]
  \centering
  \begin{tcolorbox}[title=\textbf{Rule-based Prompt for \texttt{TemplateAgent}}, colback=gray!5, colframe=gray!40!black, fontupper=\small]

    You are a CRM copywriting assistant. Please rewrite the original message below into a short, action-driven, persuasive push notification. Do not focus on personalization customization. Focus on improving clarity, structure, and incentive framing --- while preserving the original intent. Use a more engaging tone and clear incentive structure. Keep the meaning, but improve clarity and impact. Respond in the similar language style as the Original Poor Template.\\

    === Original Message ===\\
    \texttt{\{original\}}
    
  \end{tcolorbox}
  \caption{Prompt used by \texttt{TemplateAgent} for rule-based rewriting of underperforming CRM messages when no suitable exemplars are available.}
  \label{fig:prompt_template_agent_rule_based}
\end{figure}

%%%%%%%%%%%%%%%%%%%%%%%%%%%%%%
\subsection{\texttt{EvaluateAgent}}

\begin{figure}[h!]
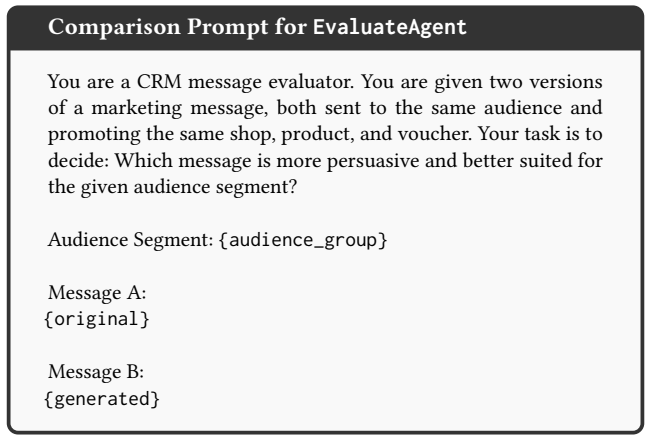

  \centering
  \begin{tcolorbox}[title=\textbf{Comparison Prompt for \texttt{EvaluateAgent}}, colback=gray!5, colframe=gray!40!black, fontupper=\small]

    You are a CRM message evaluator. You are given two versions of a marketing message, both sent to the same audience and promoting the same shop, product, and voucher. Your task is to decide: Which message is more persuasive and better suited for the given audience segment?\\

    Audience Segment: \texttt{\{audience\_group\}}\\

    Message A:\\
    \texttt{\{original\}}\\

    Message B:\\
    \texttt{\{generated\}}
  \end{tcolorbox}
  \caption{Prompt for comparative evaluation, where \texttt{EvaluateAgent} selects the more persuasive message without knowing which version is original.}
  \label{fig:prompt_evaluate_agent_2}
\end{figure}

\texttt{EvaluateAgent} assesses the quality of CRM message template revisions through two complementary evaluation modes. In the scoring mode, it rates each message independently on two criteria: (1) audience match and (2) marketing effectiveness. Each criterion uses a three-level scale with values of 1 (poor), 3 (average), and 5 (excellent). Concise justifications are provided to support each score and enhance interpretability. In the comparison mode, the agent receives two anonymized versions labeled as A and B and selects the more persuasive and audience-appropriate candidate. The agent is not told which version is original and which is rewritten, ensuring unbiased evaluation based solely on message quality. The prompt for the scoring mode appears in Figure~\ref{fig:prompt_evaluate_agent_1}, and the prompt for the comparison mode is shown in Figure~\ref{fig:prompt_evaluate_agent_2}.

\begin{figure}[h!]
  \centering
  \begin{tcolorbox}[title=\textbf{Scoring Prompt for \texttt{EvaluateAgent}}, colback=gray!5, colframe=gray!40!black, fontupper=\small]
  
    You are a CRM message evaluator. You are given two marketing messages for the same audience segment, promoting the same shop, product, and voucher. The only difference is the message content. Please evaluate both Message A and Message B based on the following two criteria:\\
    1. Audience Match --- How well does the message suit the audience segment: \texttt{\{audience\_group\}}?\\
    2. Marketing Effectiveness --- Is the message clear, persuasive, and motivating?\\

    Use a scoring scale with three levels only: 1 = poor · 3 = average · 5 = excellent.\\

    Audience Segment: \texttt{\{audience\_group\}}\\

    Message A:\\
    \texttt{\{original\}}\\

    Message B:\\
    \texttt{\{generated\}}
  \end{tcolorbox}
  \caption{Prompt used by \texttt{EvaluateAgent} to score CRM messages by audience match and marketing effectiveness.}
  \label{fig:prompt_evaluate_agent_1}
\end{figure}

%%%%%%%%%%%%%%%%%%%%%%%%%%%%%%
\section{Experiments}

We conduct comprehensive experiments to assess the effectiveness of our proposed multi-agent framework for CRM message template generation. This section is organized into three parts: we first present quantitative evaluation results across audience segments, analyzing audience alignment, marketing quality, and preference metrics. We then perform an ablation study to examine the impact of different LLM configurations for each agent component. Finally, we include qualitative case studies to highlight how generated messages differ from the originals in structure, tone, and targeting strategy.

\subsection{Experimental Results}

\begin{table*}[h!]
\small
\centering
\caption{Evaluation performance of original and generated CRM messages across audience segments. Audience score and market score are produced by \texttt{EvaluateAgent} to measure alignment with the target segment and promotional effectiveness, respectively. The Preference columns report which version (original or generated) is favored by \texttt{EvaluateAgent} in blind comparisons, along with the corresponding selection rate. $\Delta$ (\%) indicates relative improvement from original to generated messages. (Ori: Original; Gen: Generated.)}
\begin{tabular}{lrrrrrrrrrr}
\toprule
\multirow{2}{*}[-0.8ex]{\textbf{Audience Segment}} &
\multirow{2}{*}[-0.8ex]{\textbf{Count}} &
\multicolumn{3}{c}{\textbf{Audience Score}} &
\multicolumn{3}{c}{\textbf{Market Score}} &
\multicolumn{2}{c}{\textbf{Preference}} \\
\cmidrule(lr){3-5} \cmidrule(lr){6-8} \cmidrule(lr){9-10}
 & & \textbf{Ori} & \textbf{Gen} & \textbf{$\Delta$ (\%)} 
   & \textbf{Ori} & \textbf{Gen} & \textbf{$\Delta$ (\%)}
   & \textbf{Preferred} & \textbf{Rate (\%)} \\
\midrule
Potential New Customers & 701 & 3.32 & 4.71 & 41.87 & 3.14 & 4.78 & 52.23 & Gen & 88.16 \\
New Buyers              & 681 & 3.63 & 4.38 & 20.66 & 3.32 & 4.57 & 37.65 & Gen & 80.62 \\
Lapsing Buyers          & 515 & 3.84 & 4.12 &  7.29 & 3.41 & 4.38 & 28.45 & Gen & 67.18 \\
Abandoned Cart Buyers   & 493 & 4.76 & 4.91 &  3.15 & 3.17 & 4.83 & 52.37 & Gen & 87.22 \\
Unpaid Order Buyers     & 439 & 4.95 & 4.89 & -1.21 & 3.30 & 4.73 & 43.33 & Gen & 83.60 \\
Post-Purchase Group     & 352 & 4.91 & 4.74 & -3.46 & 3.18 & 4.56 & 43.40 & Gen & 76.70 \\
Price-Drop Group        & 267 & 4.99 & 4.79 & -4.01 & 4.03 & 4.39 &  8.93 & Gen & 63.67 \\
Active Old Followers    & 186 & 4.06 & 3.89 & -4.19 & 3.25 & 4.60 & 41.54 & Gen & 70.43 \\
Frequent Buyers         & 140 & 4.20 & 4.20 &  0.00 & 3.54 & 4.44 & 25.42 & Gen & 64.29 \\
New Followers           & 125 & 4.70 & 4.46 & -5.11 & 3.77 & 4.52 & 19.89 & Gen & 79.20 \\
Repeat Buyers           &  58 & 4.17 & 4.00 & -4.08 & 3.41 & 4.34 & 27.27 & Gen & 56.90 \\
\midrule
\textbf{Overall}        &  3957    & 4.18   & 4.56     &   9.09   &  3.33    &   4.61  & 38.44  & Gen & 78.44 \\
\bottomrule
\end{tabular}
\label{tab:audience_segment_summary}
\end{table*}

We evaluate the quality of CRMAgent generated templates against merchant originals across 11 audience segments, including potential new customers, new buyers, lapsing buyers, abandoned cart buyers, unpaid order buyers, post-purchase users, price-drop seekers, active old followers, frequent buyers, new followers, and repeat buyers. First, we quantify gains in audience alignment, marketing effectiveness, and overall preference using the structured scores produced by \texttt{EvaluateAgent}. Next, we compute BERTScore-F1 and chrF to verify that the rewritten copy preserves the core semantic content while allowing phrasing changes. Together, these two sets of metrics offer a comprehensive view of both persuasive impact and faithfulness to the original intent.

We use GPT-4o \citep{openai2024gpt4o} as the underlying model for \texttt{ContentAgent}, \texttt{TemplateAgent}, and \texttt{EvaluateAgent} in this experiment. Table \ref{tab:audience_segment_summary} shows that the generated templates consistently outperform the originals across most audience segments. Overall, the audience fit score rises by 9.09\%, and the marketing effectiveness score jumps by 38.44\%. \texttt{EvaluateAgent} chooses the rewritten version in 78.44\% of blind comparisons, confirming its superiority. Qualitatively, segments that typically need a stronger push, such as potential new customers, benefit the most: the new copy adds explicit urgency and a clearer value proposition, making it more compelling to first time shoppers. In contrast, segments like frequent buyers show only modest gains because the baseline messages are already well targeted; improvements here mainly come from tighter calls to action rather than drastic content changes. A few groups, for example price-drop seekers, record a slight dip in audience fit score, likely because the rewritten text trades some price sensitivity cues for broader appeal. Even so, their marketing scores still rise, indicating that the generated messages remain more persuasive overall despite minor targeting trade-offs.

To complement the evaluation of persuasive quality, we also measure how closely the generated messages preserve the meaning and style of the originals. Table~\ref{tab:semantic_metrics} reports semantic and surface-level similarity scores using BERTScore-F1 \citep{zhang2019bertscore} and chrF \citep{popovic2015chrf}, respectively. BERTScore-F1, computed with the multilingual \texttt{xlm-roberta-large} \citep{DBLP:journals/corr/abs-1911-02116} model, captures token-level semantic alignment between the original and generated messages. Meanwhile, chrF quantifies surface-level overlap based on character and word n-gram matches. The overall BERTScore-F1 of 0.87 and chrF of 24.22 suggest that while the rewrites are not mere copies, they still maintain strong semantic consistency with the source messages---allowing for effective reformulation without drifting from the original intent.

\begin{table*}[h!]
\small
\centering
\caption{Ablation study comparing different LLM configurations for each agent module. (Ori: Original; Gen: Generated.)}
\begin{tabular}{lllrrrrrrrr}
\toprule
\multirow{2}{*}[-0.8ex]{\textbf{Content}} &
\multirow{2}{*}[-0.8ex]{\textbf{Template}} &
\multirow{2}{*}[-0.8ex]{\textbf{Evaluate}} &
\multicolumn{3}{c}{\textbf{Audience Score}} &
\multicolumn{3}{c}{\textbf{Market Score}} &
\multicolumn{2}{c}{\textbf{Preference}} \\
\cmidrule(lr){4-6} \cmidrule(lr){7-9} \cmidrule(lr){10-11}
 & & & \textbf{Ori} & \textbf{Gen} & \textbf{$\Delta$ (\%)}
 & \textbf{Ori} & \textbf{Gen} & \textbf{$\Delta$ (\%)} %
 & \textbf{Preferred} & \textbf{Rate (\%)} \\
\midrule
GPT-4o & GPT-4o           & GPT-4o            & 4.18   & 4.56     &   9.09   &  3.33    &   4.61  & 38.44  & Gen & 78.44 \\
GPT-4o & GPT-4o + CoT     & GPT-4o            &  4.27     &   4.26    &   -0.23     &   3.45    &    4.44   &    28.70   & Gen & 61.51    \\
GPT-4o & GPT-4o           & GPT-4o + CoT      &   3.80    &   4.30    &    13.16   &    3.25   &    4.49   &   38.15    & Gen & 81.17    \\
GPT-4o & GPT-4.1 mini     & GPT-4o            &   4.19    &    4.54   &    8.35   &    3.32   &   4.60    &   38.55    & Gen & 78.47    \\
GPT-4o & GPT-4o           & GPT-4.1 mini      &   4.20    &    4.43   &    5.48   &    3.30   &    4.61   &   39.70    & Gen & 81.05    \\
GPT-4o & GPT-4o           & DeepSeek-R1       &   3.39    &    4.51   &    33.04   &   3.08    &   4.84    &   57.14    & Gen & 55.40    \\
\bottomrule
\end{tabular}
\label{tab:ablation_study}
\end{table*}

\begin{table}[h!]
\small
\centering
\caption{Similarity between original and generated CRM messages across audience segments, measured by BERTScore-F1 and chrF.}
\begin{tabular}{lrr}
\toprule
\textbf{Audience Segment} & \textbf{BERTScore-F1} & \textbf{chrF} \\
\midrule
Potential New Customers     & 0.87 & 23.44 \\
New Buyers                  & 0.87 & 23.15 \\
Lapsing Buyers              & 0.87 & 21.24 \\
Abandoned Cart Buyers       & 0.87 & 25.26 \\
Unpaid Order Buyers         & 0.87 & 25.30 \\
Post-Purchase Group         & 0.89 & 30.67 \\
Price-Drop Group            & 0.89 & 26.13 \\
Active Old Followers        & 0.87 & 20.48 \\
Frequent Buyers             & 0.86 & 24.49 \\
New Followers               & 0.88 & 22.80 \\
Repeat Buyers               & 0.86 & 22.46 \\
\midrule
\textbf{Overall}            & 0.87 & 24.22 \\
\bottomrule
\end{tabular}
\label{tab:semantic_metrics}
\end{table}

%%%%%%%%%%%%%%%%%%%%%%%%%%%%%%
\subsection{Ablation Studies}

To assess how different LLM configurations affect system performance, we conduct an ablation study by varying the backbone models used in \texttt{TemplateAgent} and \texttt{EvaluateAgent}. Specifically, we explore combinations involving GPT-4o \citep{openai2024gpt4o}, GPT-4.1 mini \citep{openai2025gpt41}, and DeepSeek-R1 \citep{guo2025deepseek}, as well as the use of chain-of-thought (CoT) \citep{wei2022chain} prompting. \texttt{ContentAgent} is kept fixed as GPT-4o to isolate the effects of template generation and evaluation components.

Table~\ref{tab:ablation_study} shows that while the fully GPT-4o configuration delivers strong overall results, it is not uniformly superior. For example, applying chain-of-thought prompting to \texttt{EvaluateAgent} slightly boosts preference selection compared to the baseline, suggesting benefits in evaluation reasoning. In contrast, using CoT in \texttt{TemplateAgent} leads to degraded outcomes, likely due to verbose or unfocused generation. Models like GPT-4.1 mini offer performance comparable to GPT-4o, achieving similarly strong improvements in both audience and market scores, as well as high preference rates. This suggests that lighter-weight models can serve as viable drop-in replacements when resource constraints are a concern. In contrast, DeepSeek-R1 as \texttt{EvaluateAgent} yields the highest score improvements, but its relatively low preference rate indicates a potential mismatch between its scoring criteria and human-aligned quality judgments.

%%%%%%%%%%%%%%%%%%%%%%%%%%%%%%
\subsection{Case Studies}

To illustrate how the CRMAgent adapts its messaging strategy across audience types, Table~\ref{tab:case_study} presents two representative examples from potential new customers and abandoned cart buyers segments. In the first case, the original message is generic and celebratory, while the generated version introduces urgency (``Limited Time Offer'') and a clearer call to action, better aligning with the motivational needs of first-time shoppers. In the second example, the original offers a mild reminder, but the revised message adds a direct discount (``RM3 off'') and an upsell condition, using persuasive language to prompt checkout completion. These cases highlight how the agent enhances promotional effectiveness through more targeted and action-oriented rewrites.

\begin{table}[h!]
\small
\centering
\caption{Illustrative case study comparing original and generated CRM message templates for the potential new customers and abandoned cart buyers segments.}
\begin{tabular}{p{0.9\linewidth}}
\toprule
\textbf{Case Study 1: Potential New Customers} \\ % 12467
\midrule
\textbf{Original:}  Everyone's loving these! \\
Shoppers love these best-sellers---find your next favorite in our latest collection. \\

\textbf{Generated:}  Limited Time Offer! \\
Don't miss out on our hottest best-sellers! Explore our latest collection now and be the first to secure your favorites before they're gone. Shop today! \\
\midrule
\textbf{Case Study 2: Abandoned Cart Buyers} \\ % 1667
\midrule
\textbf{Original:}  Did you forget something? \\
Looks like you left something in your cart. Ready to pick up where you left off? \\

\textbf{Generated:}  Your cart is waiting---grab RM3 off now! \\
You left great picks in your cart! Complete your purchase with RM3 off when you spend RM15 or more. Don't miss out---shop now and enjoy the benefits! \\
\bottomrule
\end{tabular}
% \caption{Illustrative case study comparing original and generated CRM message templates for the potential new customers and abandoned cart buyers segments.}
\label{tab:case_study}
\end{table}

%%%%%%%%%%%%%%%%%%%%%%%%%%%%%%
\section{Conclusion}

In this paper, we introduce CRMAgent, a multi-agent system powered by LLMs for generating high-quality CRM message templates in the e-commerce domain. The system integrates three complementary strategies: content diagnosis, retrieval based adaptation, and rule based fallback to support merchants in crafting action oriented, persuasive copy across diverse audience segments. Extensive experiments show that CRMAgent consistently improves both audience alignment and marketing effectiveness compared to merchant written messages. In blind evaluations, the generated messages are preferred in the majority of cases, demonstrating the practical impact of our system. These findings suggest that multi-agent LLM systems can serve as scalable and effective solutions for automated CRM content creation, with potential to support merchants of varying expertise levels.

% \begin{acks}
% \end{acks}

\bibliographystyle{ACM-Reference-Format}
\bibliography{sample-base}


\begin{thebibliography}{29}

%%% the Web address.

\ifx \showCODEN    \undefined \def \showCODEN     #1{\unskip}     \fi
\ifx \showISBNx    \undefined \def \showISBNx     #1{\unskip}     \fi
\ifx \showISBNxiii \undefined \def \showISBNxiii  #1{\unskip}     \fi
\ifx \showISSN     \undefined \def \showISSN      #1{\unskip}     \fi
\ifx \showLCCN     \undefined \def \showLCCN      #1{\unskip}     \fi
\ifx \shownote     \undefined \def \shownote      #1{#1}          \fi
\ifx \showarticletitle \undefined \def \showarticletitle #1{#1}   \fi
\ifx \showURL      \undefined \def \showURL       {\relax}        \fi
% invisible to TeX
\providecommand\bibfield[2]{#2}
\providecommand\bibinfo[2]{#2}
\providecommand\natexlab[1]{#1}
\providecommand\showeprint[2][]{arXiv:#2}

\bibitem[Chen et~al\mbox{.}(2024)]%
        {chen2024knowledge}
\bibfield{author}{\bibinfo{person}{Yuemin Chen}, \bibinfo{person}{Feifan Wu}, \bibinfo{person}{Jingwei Wang}, \bibinfo{person}{Hao Qian}, \bibinfo{person}{Ziqi Liu}, \bibinfo{person}{Zhiqiang Zhang}, \bibinfo{person}{Jun Zhou}, {and} \bibinfo{person}{Meng Wang}.} \bibinfo{year}{2024}\natexlab{}.
\newblock \showarticletitle{Knowledge-augmented Financial Market Analysis and Report Generation}. In \bibinfo{booktitle}{\emph{Proceedings of the 2024 Conference on Empirical Methods in Natural Language Processing: Industry Track}}. \bibinfo{pages}{1207--1217}.
\newblock


\bibitem[Conneau et~al\mbox{.}(2019)]%
        {DBLP:journals/corr/abs-1911-02116}
\bibfield{author}{\bibinfo{person}{Alexis Conneau}, \bibinfo{person}{Kartikay Khandelwal}, \bibinfo{person}{Naman Goyal}, \bibinfo{person}{Vishrav Chaudhary}, \bibinfo{person}{Guillaume Wenzek}, \bibinfo{person}{Francisco Guzm{\'{a}}n}, \bibinfo{person}{Edouard Grave}, \bibinfo{person}{Myle Ott}, \bibinfo{person}{Luke Zettlemoyer}, {and} \bibinfo{person}{Veselin Stoyanov}.} \bibinfo{year}{2019}\natexlab{}.
\newblock \showarticletitle{Unsupervised Cross-lingual Representation Learning at Scale}.
\newblock \bibinfo{journal}{\emph{CoRR}}  \bibinfo{volume}{abs/1911.02116} (\bibinfo{year}{2019}).
\newblock
\showeprint[arXiv]{1911.02116}
\urldef\tempurl%
\url{http://arxiv.org/abs/1911.02116}
\showURL{%
\tempurl}


\bibitem[Douze et~al\mbox{.}(2024)]%
        {douze2024faiss}
\bibfield{author}{\bibinfo{person}{Matthijs Douze}, \bibinfo{person}{Alexandr Guzhva}, \bibinfo{person}{Chengqi Deng}, \bibinfo{person}{Jeff Johnson}, \bibinfo{person}{Gergely Szilvasy}, \bibinfo{person}{Pierre-Emmanuel Mazaré}, \bibinfo{person}{Maria Lomeli}, \bibinfo{person}{Lucas Hosseini}, {and} \bibinfo{person}{Hervé Jégou}.} \bibinfo{year}{2024}\natexlab{}.
\newblock \showarticletitle{The Faiss library}.
\newblock  (\bibinfo{year}{2024}).
\newblock
\showeprint[arxiv]{2401.08281}~[cs.LG]


\bibitem[Guo et~al\mbox{.}(2025)]%
        {guo2025deepseek}
\bibfield{author}{\bibinfo{person}{Daya Guo}, \bibinfo{person}{Dejian Yang}, \bibinfo{person}{Haowei Zhang}, \bibinfo{person}{Junxiao Song}, \bibinfo{person}{Ruoyu Zhang}, \bibinfo{person}{Runxin Xu}, \bibinfo{person}{Qihao Zhu}, \bibinfo{person}{Shirong Ma}, \bibinfo{person}{Peiyi Wang}, \bibinfo{person}{Xiao Bi}, {et~al\mbox{.}}} \bibinfo{year}{2025}\natexlab{}.
\newblock \showarticletitle{Deepseek-r1: Incentivizing reasoning capability in llms via reinforcement learning}.
\newblock \bibinfo{journal}{\emph{arXiv preprint arXiv:2501.12948}} (\bibinfo{year}{2025}).
\newblock


\bibitem[Guo and Yang(2024)]%
        {guo2024econnli}
\bibfield{author}{\bibinfo{person}{Yue Guo} {and} \bibinfo{person}{Yi Yang}.} \bibinfo{year}{2024}\natexlab{}.
\newblock \showarticletitle{EconNLI: Evaluating Large Language Models on Economics Reasoning}.
\newblock \bibinfo{journal}{\emph{arXiv preprint arXiv:2407.01212}} (\bibinfo{year}{2024}).
\newblock


\bibitem[Hashimoto et~al\mbox{.}(2018)]%
        {hashimoto2018retrieve}
\bibfield{author}{\bibinfo{person}{Tatsunori~B Hashimoto}, \bibinfo{person}{Kelvin Guu}, \bibinfo{person}{Yonatan Oren}, {and} \bibinfo{person}{Percy~S Liang}.} \bibinfo{year}{2018}\natexlab{}.
\newblock \showarticletitle{A retrieve-and-edit framework for predicting structured outputs}.
\newblock \bibinfo{journal}{\emph{Advances in Neural Information Processing Systems}}  \bibinfo{volume}{31} (\bibinfo{year}{2018}).
\newblock


\bibitem[Ieva et~al\mbox{.}(2025)]%
        {ieva2025enhancing}
\bibfield{author}{\bibinfo{person}{Saverio Ieva}, \bibinfo{person}{Ivano Bilenchi}, \bibinfo{person}{Filippo Gramegna}, \bibinfo{person}{Agnese Pinto}, \bibinfo{person}{Floriano Scioscia}, \bibinfo{person}{Michele Ruta}, {and} \bibinfo{person}{Giuseppe Loseto}.} \bibinfo{year}{2025}\natexlab{}.
\newblock \showarticletitle{Enhancing Last-Mile Logistics: AI-Driven Fleet Optimization, Mixed Reality, and Large Language Model Assistants for Warehouse Operations}.
\newblock \bibinfo{journal}{\emph{Sensors}} \bibinfo{volume}{25}, \bibinfo{number}{9} (\bibinfo{year}{2025}), \bibinfo{pages}{2696}.
\newblock


\bibitem[Kumar and Reinartz(2018)]%
        {kumar2018customer}
\bibfield{author}{\bibinfo{person}{Vineet Kumar} {and} \bibinfo{person}{Werner Reinartz}.} \bibinfo{year}{2018}\natexlab{}.
\newblock \bibinfo{booktitle}{\emph{Customer relationship management}}.
\newblock \bibinfo{publisher}{Springer}.
\newblock


\bibitem[Li et~al\mbox{.}(2023)]%
        {li2023large}
\bibfield{author}{\bibinfo{person}{Beibin Li}, \bibinfo{person}{Konstantina Mellou}, \bibinfo{person}{Bo Zhang}, \bibinfo{person}{Jeevan Pathuri}, {and} \bibinfo{person}{Ishai Menache}.} \bibinfo{year}{2023}\natexlab{}.
\newblock \showarticletitle{Large language models for supply chain optimization}.
\newblock \bibinfo{journal}{\emph{arXiv preprint arXiv:2307.03875}} (\bibinfo{year}{2023}).
\newblock


\bibitem[Liu et~al\mbox{.}(2024)]%
        {liu2024fine}
\bibfield{author}{\bibinfo{person}{Yuhang Liu}, \bibinfo{person}{Xueyu Hu}, \bibinfo{person}{Shengyu Zhang}, \bibinfo{person}{Jingyuan Chen}, \bibinfo{person}{Fan Wu}, {and} \bibinfo{person}{Fei Wu}.} \bibinfo{year}{2024}\natexlab{}.
\newblock \showarticletitle{Fine-Grained Guidance for Retrievers: Leveraging LLMs' Feedback in Retrieval-Augmented Generation}.
\newblock \bibinfo{journal}{\emph{arXiv preprint arXiv:2411.03957}} (\bibinfo{year}{2024}).
\newblock


\bibitem[Liu and Quan(2024)]%
        {liu2024tpp}
\bibfield{author}{\bibinfo{person}{Zefang Liu} {and} \bibinfo{person}{Yinzhu Quan}.} \bibinfo{year}{2024}\natexlab{}.
\newblock \showarticletitle{TPP-LLM: Modeling Temporal Point Processes by Efficiently Fine-Tuning Large Language Models}.
\newblock \bibinfo{journal}{\emph{arXiv preprint arXiv:2410.02062}} (\bibinfo{year}{2024}).
\newblock


\bibitem[Liu and Quan(2025a)]%
        {liu2025econwebarena}
\bibfield{author}{\bibinfo{person}{Zefang Liu} {and} \bibinfo{person}{Yinzhu Quan}.} \bibinfo{year}{2025}\natexlab{a}.
\newblock \showarticletitle{EconWebArena: Benchmarking Autonomous Agents on Economic Tasks in Realistic Web Environments}.
\newblock \bibinfo{journal}{\emph{arXiv preprint arXiv:2506.08136}} (\bibinfo{year}{2025}).
\newblock


\bibitem[Liu and Quan(2025b)]%
        {liu2025retrieval}
\bibfield{author}{\bibinfo{person}{Zefang Liu} {and} \bibinfo{person}{Yinzhu Quan}.} \bibinfo{year}{2025}\natexlab{b}.
\newblock \showarticletitle{Retrieval of temporal event sequences from textual descriptions}. In \bibinfo{booktitle}{\emph{Proceedings of the 4th International Workshop on Knowledge-Augmented Methods for Natural Language Processing}}. \bibinfo{pages}{37--49}.
\newblock


\bibitem[Malik et~al\mbox{.}(2024)]%
        {malik2024pearl}
\bibfield{author}{\bibinfo{person}{Vijit Malik}, \bibinfo{person}{Akshay Jagatap}, \bibinfo{person}{Vinayak Puranik}, {and} \bibinfo{person}{Anirban Majumder}.} \bibinfo{year}{2024}\natexlab{}.
\newblock \showarticletitle{PEARL: Preference extraction with exemplar augmentation and retrieval with LLM agents}. In \bibinfo{booktitle}{\emph{Proceedings of the 2024 Conference on Empirical Methods in Natural Language Processing: Industry Track}}. \bibinfo{pages}{1536--1547}.
\newblock


\bibitem[{OpenAI}(2024)]%
        {openai2024gpt4o}
\bibfield{author}{\bibinfo{person}{{OpenAI}}.} \bibinfo{year}{2024}\natexlab{}.
\newblock \bibinfo{title}{Hello GPT-4o}.
\newblock
\urldef\tempurl%
\url{https://openai.com/index/hello-gpt-4o/}
\showURL{%
\tempurl}


\bibitem[{OpenAI}(2025a)]%
        {openai2025gpt41}
\bibfield{author}{\bibinfo{person}{{OpenAI}}.} \bibinfo{year}{2025}\natexlab{a}.
\newblock \bibinfo{title}{Introducing GPT-4.1 in the API}.
\newblock
\urldef\tempurl%
\url{https://openai.com/index/gpt-4-1/}
\showURL{%
\tempurl}


\bibitem[{OpenAI}(2025b)]%
        {openai2025o4mini}
\bibfield{author}{\bibinfo{person}{{OpenAI}}.} \bibinfo{year}{2025}\natexlab{b}.
\newblock \bibinfo{title}{Introducing OpenAI o3 and o4-mini}.
\newblock
\urldef\tempurl%
\url{https://openai.com/index/introducing-o3-and-o4-mini/}
\showURL{%
\tempurl}


\bibitem[Popovi{\'c}(2015)]%
        {popovic2015chrf}
\bibfield{author}{\bibinfo{person}{Maja Popovi{\'c}}.} \bibinfo{year}{2015}\natexlab{}.
\newblock \showarticletitle{chrF: character n-gram F-score for automatic MT evaluation}. In \bibinfo{booktitle}{\emph{Proceedings of the tenth workshop on statistical machine translation}}. \bibinfo{pages}{392--395}.
\newblock


\bibitem[Quan and Liu(2024a)]%
        {quan2024econlogicqa}
\bibfield{author}{\bibinfo{person}{Yinzhu Quan} {and} \bibinfo{person}{Zefang Liu}.} \bibinfo{year}{2024}\natexlab{a}.
\newblock \showarticletitle{Econlogicqa: A question-answering benchmark for evaluating large language models in economic sequential reasoning}.
\newblock \bibinfo{journal}{\emph{arXiv preprint arXiv:2405.07938}} (\bibinfo{year}{2024}).
\newblock


\bibitem[Quan and Liu(2024b)]%
        {quan2024invagent}
\bibfield{author}{\bibinfo{person}{Yinzhu Quan} {and} \bibinfo{person}{Zefang Liu}.} \bibinfo{year}{2024}\natexlab{b}.
\newblock \showarticletitle{Invagent: A large language model based multi-agent system for inventory management in supply chains}.
\newblock \bibinfo{journal}{\emph{arXiv preprint arXiv:2407.11384}} (\bibinfo{year}{2024}).
\newblock


\bibitem[Quan et~al\mbox{.}(2025)]%
        {quan2025leveraging}
\bibfield{author}{\bibinfo{person}{Yinzhu Quan}, \bibinfo{person}{Yujia Xu}, \bibinfo{person}{Guanlin Chen}, \bibinfo{person}{Frederick Benaben}, {and} \bibinfo{person}{Benoit Montreuil}.} \bibinfo{year}{2025}\natexlab{}.
\newblock \showarticletitle{Leveraging Large Language Models for Risk Assessment in Hyperconnected Logistic Hub Network Deployment}.
\newblock \bibinfo{journal}{\emph{arXiv preprint arXiv:2503.21115}} (\bibinfo{year}{2025}).
\newblock


\bibitem[Reimers and Gurevych(2019)]%
        {reimers2019sentence}
\bibfield{author}{\bibinfo{person}{Nils Reimers} {and} \bibinfo{person}{Iryna Gurevych}.} \bibinfo{year}{2019}\natexlab{}.
\newblock \showarticletitle{Sentence-bert: Sentence embeddings using siamese bert-networks}.
\newblock \bibinfo{journal}{\emph{arXiv preprint arXiv:1908.10084}} (\bibinfo{year}{2019}).
\newblock


\bibitem[Singh et~al\mbox{.}(2024)]%
        {singh2024personal}
\bibfield{author}{\bibinfo{person}{Harmanpreet Singh}, \bibinfo{person}{Nikhil Verma}, \bibinfo{person}{Yixiao Wang}, \bibinfo{person}{Manasa Bharadwaj}, \bibinfo{person}{Homa Fashandi}, \bibinfo{person}{Kevin Ferreira}, {and} \bibinfo{person}{Chul Lee}.} \bibinfo{year}{2024}\natexlab{}.
\newblock \showarticletitle{Personal Large Language Model Agents: A Case Study on Tailored Travel Planning}. In \bibinfo{booktitle}{\emph{Proceedings of the 2024 Conference on Empirical Methods in Natural Language Processing: Industry Track}}. \bibinfo{pages}{486--514}.
\newblock


\bibitem[Wang et~al\mbox{.}(2025)]%
        {wang2025large}
\bibfield{author}{\bibinfo{person}{Shenao Wang}, \bibinfo{person}{Yanjie Zhao}, \bibinfo{person}{Xinyi Hou}, {and} \bibinfo{person}{Haoyu Wang}.} \bibinfo{year}{2025}\natexlab{}.
\newblock \showarticletitle{Large language model supply chain: A research agenda}.
\newblock \bibinfo{journal}{\emph{ACM Transactions on Software Engineering and Methodology}} \bibinfo{volume}{34}, \bibinfo{number}{5} (\bibinfo{year}{2025}), \bibinfo{pages}{1--46}.
\newblock


\bibitem[Wang et~al\mbox{.}(2020)]%
        {wang2020minilm}
\bibfield{author}{\bibinfo{person}{Wenhui Wang}, \bibinfo{person}{Furu Wei}, \bibinfo{person}{Li Dong}, \bibinfo{person}{Hangbo Bao}, \bibinfo{person}{Nan Yang}, {and} \bibinfo{person}{Ming Zhou}.} \bibinfo{year}{2020}\natexlab{}.
\newblock \showarticletitle{Minilm: Deep self-attention distillation for task-agnostic compression of pre-trained transformers}.
\newblock \bibinfo{journal}{\emph{Advances in neural information processing systems}}  \bibinfo{volume}{33} (\bibinfo{year}{2020}), \bibinfo{pages}{5776--5788}.
\newblock


\bibitem[Wei et~al\mbox{.}(2022)]%
        {wei2022chain}
\bibfield{author}{\bibinfo{person}{Jason Wei}, \bibinfo{person}{Xuezhi Wang}, \bibinfo{person}{Dale Schuurmans}, \bibinfo{person}{Maarten Bosma}, \bibinfo{person}{Fei Xia}, \bibinfo{person}{Ed Chi}, \bibinfo{person}{Quoc~V Le}, \bibinfo{person}{Denny Zhou}, {et~al\mbox{.}}} \bibinfo{year}{2022}\natexlab{}.
\newblock \showarticletitle{Chain-of-thought prompting elicits reasoning in large language models}.
\newblock \bibinfo{journal}{\emph{Advances in neural information processing systems}}  \bibinfo{volume}{35} (\bibinfo{year}{2022}), \bibinfo{pages}{24824--24837}.
\newblock


\bibitem[Yang et~al\mbox{.}(2025)]%
        {yang2025docagent}
\bibfield{author}{\bibinfo{person}{Dayu Yang}, \bibinfo{person}{Antoine Simoulin}, \bibinfo{person}{Xin Qian}, \bibinfo{person}{Xiaoyi Liu}, \bibinfo{person}{Yuwei Cao}, \bibinfo{person}{Zhaopu Teng}, {and} \bibinfo{person}{Grey Yang}.} \bibinfo{year}{2025}\natexlab{}.
\newblock \showarticletitle{DocAgent: A Multi-Agent System for Automated Code Documentation Generation}.
\newblock \bibinfo{journal}{\emph{arXiv preprint arXiv:2504.08725}} (\bibinfo{year}{2025}).
\newblock


\bibitem[Yuan et~al\mbox{.}(2024)]%
        {yuan2024maxprototyper}
\bibfield{author}{\bibinfo{person}{Mingyue Yuan}, \bibinfo{person}{Jieshan Chen}, {and} \bibinfo{person}{Aaron Quigley}.} \bibinfo{year}{2024}\natexlab{}.
\newblock \showarticletitle{MAxPrototyper: A Multi-Agent Generation System for Interactive User Interface Prototyping}.
\newblock \bibinfo{journal}{\emph{arXiv preprint arXiv:2405.07131}} (\bibinfo{year}{2024}).
\newblock


\bibitem[Zhang et~al\mbox{.}(2019)]%
        {zhang2019bertscore}
\bibfield{author}{\bibinfo{person}{Tianyi Zhang}, \bibinfo{person}{Varsha Kishore}, \bibinfo{person}{Felix Wu}, \bibinfo{person}{Kilian~Q Weinberger}, {and} \bibinfo{person}{Yoav Artzi}.} \bibinfo{year}{2019}\natexlab{}.
\newblock \showarticletitle{Bertscore: Evaluating text generation with bert}.
\newblock \bibinfo{journal}{\emph{arXiv preprint arXiv:1904.09675}} (\bibinfo{year}{2019}).
\newblock


\end{thebibliography}

\appendix

%%%%%%%%%%%%%%%%%%%%%%%%%%%%%%
\section{Data}

We begin with over 3 million successfully delivered CRM message logs from April 2025, progressively enriching each record with shop attributes, campaign metadata, product bindings, voucher information, and 7-day user engagement signals. This results in a detailed dataset indexed by plan, user, product, voucher, and template. We then group the data by key dimensions---including merchant, template, and voucher---to compute the average engagement score and sample count for each combination. In total, we obtain 15,806 aggregated records for downstream analysis.

\begin{figure}[h!]
    \centering
    \includegraphics[width=1\linewidth]{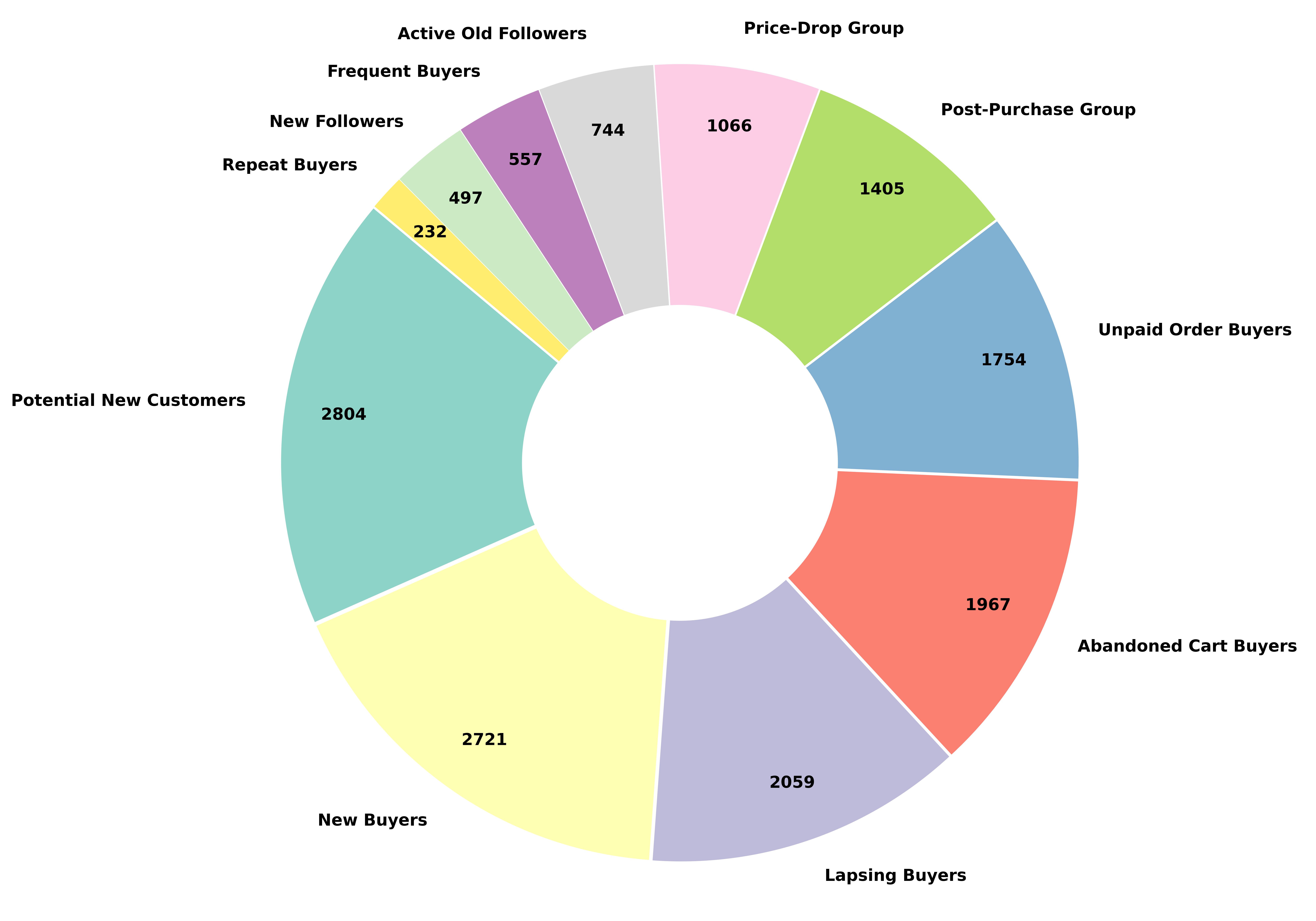}
    \caption{Distribution of CRM audience segments based on plan targeting.}
    \label{fig:crm_donut_chart}
\end{figure}

Figure~\ref{fig:crm_donut_chart} illustrates the distribution of CRM plan volume across 11 predefined audience segments, each representing a distinct stage in the customer lifecycle. For example, ``Potential New Customers'' and ``New Buyers'' account for the largest shares, suggesting merchants often focus on early-stage acquisition. In contrast, groups such as ``Repeat Buyers'' and ``Frequent Buyers'' are comparatively smaller, indicating fewer efforts are directed toward re-engagement or loyalty building. This segmentation provides essential context for interpreting downstream engagement metrics, as user behavior and responsiveness tend to vary significantly across audience types.
This breakdown helps ensure that engagement patterns are interpreted within the right audience context, rather than aggregating over inherently different customer types.

\begin{figure}[h!]
    \centering
    \includegraphics[width=1\linewidth]{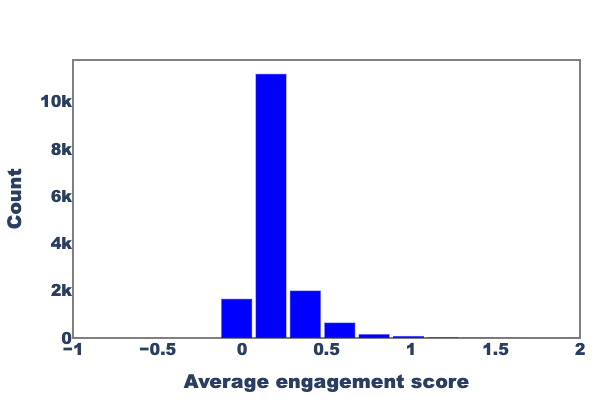}
    \caption{Distribution of average engagement scores across CRM message template plans.}
    \label{fig:distribution}
\end{figure}

The engagement score behind Figure~\ref{fig:distribution} reflects weighted user behaviors within seven days of receiving a message: one point for reading, three for clicking on a store or voucher, four for clicking a product card, two for interacting with CRM buttons, and minus five for unsubscribing. We compute this score for each unique combination of merchant, template, product, audience segment, and voucher setup. The resulting distribution is highly skewed---most plans yield low engagement, while a small number achieve high scores. Since engagement levels vary significantly across audience segments (e.g., abandoned-cart users tend to be more responsive than old followers), we analyze each segment separately. Within each group, we define the top 25\% of campaigns as strong performers and the bottom 25\% as weak ones---these serve as contrasting examples for downstream optimization.

\begin{figure}[h!]
    \centering
    \includegraphics[width=1\linewidth]{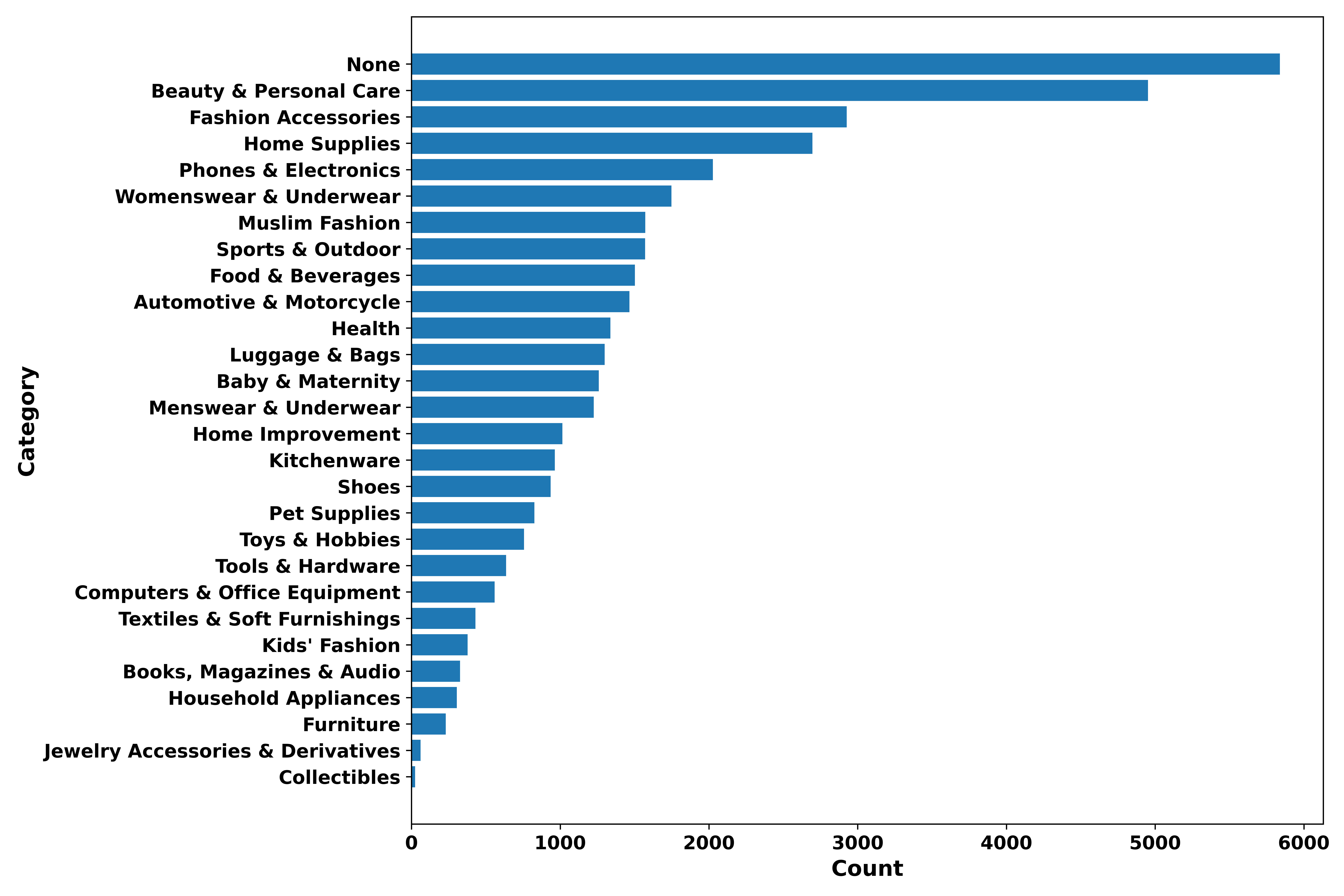}
    \caption{Distribution of CRM plans across product categories.}
    \label{fig:product_category}
\end{figure}

Figure~\ref{fig:product_category} shows the distribution of CRM plans across product categories. Many plans either lack a product binding (``None'') or focus on fast-moving goods like Beauty \& Personal Care or Fashion Accessories. This category distribution influences not only engagement patterns but also determines how templates are generated. For instance, when no product is specified, the system falls back to rule-based generation during retrieval. In contrast, when product categories are available, we can use category-matching strategies. Thus, product type plays a central role in routing template generation.

%%%%%%%%%%%%%%%%%%%%%%%%%%%%%%
\section{Agent Responses}

In this section, we present illustrative examples to demonstrate how \texttt{TemplateAgent} and \texttt{EvaluateAgent} operate. The \texttt{TemplateAgent} generates improved versions of underperforming CRM messages by adjusting tone, structure, and content to better align with audience expectations. The \texttt{EvaluateAgent} then compares the original and rewritten messages, scoring each in terms of audience match and marketing effectiveness, and selecting the preferred one.

%%%%%%%%%%%%%%%%%%%%%%%%%%%%%%
\subsection{\texttt{TemplateAgent} Responses}

\begin{figure}[h!]
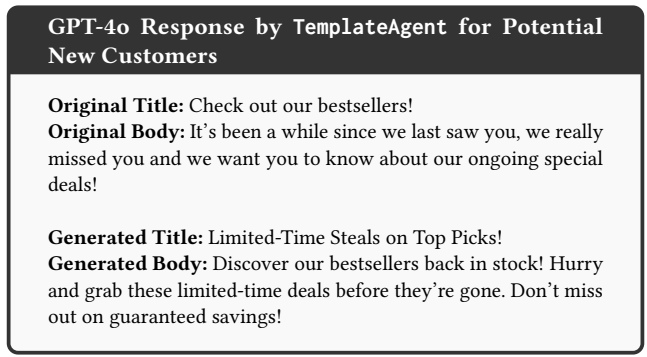

  \centering
  \begin{tcolorbox}[title=\textbf{GPT-4o Response by \texttt{TemplateAgent} for Potential New Customers}, colback=gray!5, colframe=gray!40!black, fontupper=\small]

    \textbf{Original Title:} Check out our bestsellers!\\
    \textbf{Original Body:} It's been a while since we last saw you, we really missed you and we want you to know about our ongoing special deals!\\

    \textbf{Generated Title:} Limited-Time Steals on Top Picks!\\
    \textbf{Generated Body:} Discover our bestsellers back in stock! Hurry and grab these limited-time deals before they're gone. Don't miss out on guaranteed savings!
    
  \end{tcolorbox}
  \caption{Illustrative example of \texttt{TemplateAgent} response for the potential new customer audience segment.}
  \label{fig:template_agent_example_1}
\end{figure}

\begin{figure}[h!]
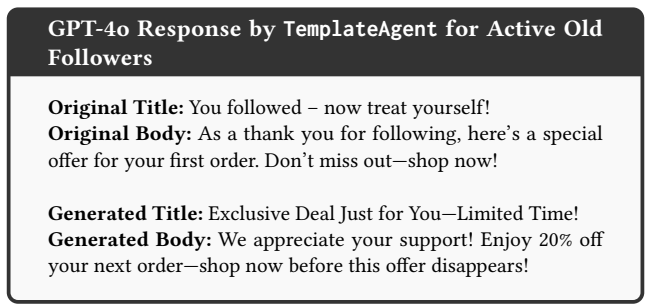

  \centering
  \begin{tcolorbox}[title=\textbf{GPT-4o Response by \texttt{TemplateAgent} for Active Old Followers}, colback=gray!5, colframe=gray!40!black, fontupper=\small]

    \textbf{Original Title:} You followed – now treat yourself!\\
    \textbf{Original Body:} As a thank you for following, here's a special offer for your first order. Don't miss out---shop now!\\

    \textbf{Generated Title:} Exclusive Deal Just for You---Limited Time!\\
    \textbf{Generated Body:} We appreciate your support! Enjoy 20\% off your next order---shop now before this offer disappears!
    
  \end{tcolorbox}
  \caption{Illustrative example of \texttt{TemplateAgent} response for the active old followers audience segment.}
  \label{fig:template_agent_example_2}
\end{figure}

To better understand how the \texttt{TemplateAgent} improves CRM messaging, we rewrite all message templates associated with poor-performing campaigns, as identified within each audience segment. Figures~\ref{fig:template_agent_example_1} and~\ref{fig:template_agent_example_2} illustrate representative examples of these rewrites for two specific audience types.

Figure~\ref{fig:template_agent_example_1} targets potential new customers who have not yet established a strong relationship with the brand and often require clear incentives to engage. The original message uses a soft, emotional tone (``we really missed you'') that, while friendly, lacks specificity and urgency. Such messaging may fail to capture the attention of users who are still undecided or unaware of the brand's value proposition. In contrast, the rewritten version is more strategic: it introduces urgency with phrases like ``limited-time steals'' and emphasizes concrete benefits such as ``guaranteed savings''. This not only strengthens the call to action but also aligns with the psychological triggers that are more effective for acquisition-focused messaging. The revised copy is concise, benefit-driven, and better tailored to convert first-time buyers by highlighting what they stand to gain immediately.

% INDEX_51

Figure~\ref{fig:template_agent_example_2} showcases a message tailored to active old followers, a segment that already has a relationship with the brand and tends to be more responsive to loyalty-oriented communication. The original message offers a vague incentive by mentioning ``a special offer for your first order'' and closes with a generic ``shop now'' prompt, which may feel impersonal or insufficient for users who have previously engaged. In contrast, the rewritten message opens with an explicit expression of appreciation and provides a clear benefit---twenty percent off the next order. This wording reinforces the value of continued engagement while making the reward tangible. The closing line, ``before this offer disappears,'' introduces urgency in a natural way, encouraging immediate action without sounding overly pushy. Overall, the revised message better aligns with the expectations and motivations of returning followers by combining gratitude, specificity, and timing.

% INDEX_2514

%%%%%%%%%%%%%%%%%%%%%%%%%%%%%%
\subsection{\texttt{EvaluateAgent} Responses}

To prevent agents from being biased by labels such as ``original'' or ``generated'', each pair of CRM messages is anonymized as Message A and Message B during evaluation. While A and B correspond to the original and generated versions respectively, the evaluators are unaware of which is which---ensuring judgments focus solely on content quality.

\begin{figure}[h!]
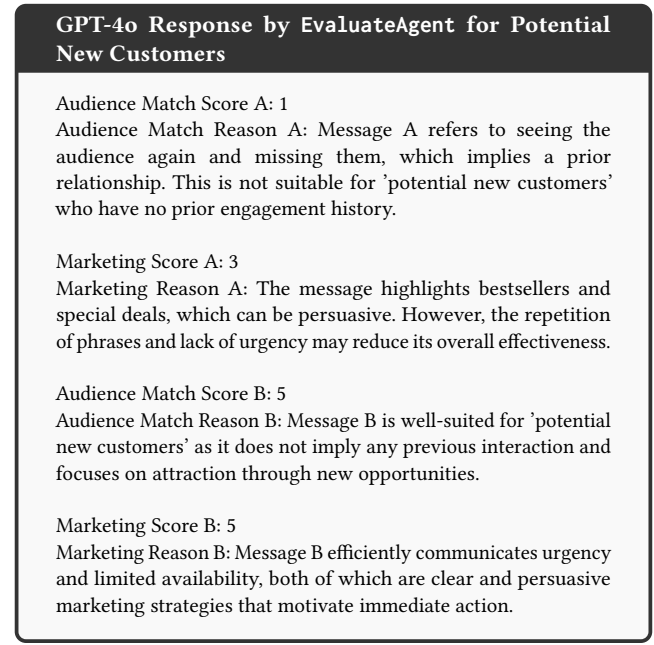

  \centering
  \begin{tcolorbox}[title=\textbf{GPT-4o Response by \texttt{EvaluateAgent} for Potential New Customers}, colback=gray!5, colframe=gray!40!black, fontupper=\small]

    Audience Match Score A: 1 \\
    Audience Match Reason A: Message A refers to seeing the audience again and missing them, which implies a prior relationship. This is not suitable for 'potential new customers' who have no prior engagement history.\\

    Marketing Score A: 3 \\
    Marketing Reason A: The message highlights bestsellers and special deals, which can be persuasive. However, the repetition of phrases and lack of urgency may reduce its overall effectiveness.\\

    Audience Match Score B: 5\\
    Audience Match Reason B: Message B is well-suited for 'potential new customers' as it does not imply any previous interaction and focuses on attraction through new opportunities.\\

    Marketing Score B: 5\\
    Marketing Reason B: Message B efficiently communicates urgency and limited availability, both of which are clear and persuasive marketing strategies that motivate immediate action.
    
  \end{tcolorbox}
  \caption{Illustrative example of \texttt{EvaluateAgent} response (audience match and marketing scores) for the potential new customer audience segment.}
  \label{fig:output_evaluate_agent_new_1}
\end{figure}

For the potential new customer segment, Figure~\ref{fig:output_evaluate_agent_new_1} illustrates a clear advantage for Message B in both audience relevance and marketing effectiveness. Message A assumes a prior relationship (``we really missed you''), which is inappropriate for new users. In contrast, Message B makes no such assumption and instead uses urgency (``limited availability'') and value (``guaranteed savings'') to attract first-time buyers. As a result, Message B receives the highest possible scores for both alignment and persuasion.

% INDEX_51

Similarly, in the active old followers segment shown in Figure~\ref{fig:output_evaluate_agent_old_1}, Message A misfires by offering a ``first order'' discount to users who are likely repeat customers. This mismatch undermines both credibility and clarity. Message B, on the other hand, acknowledges prior engagement and offers a concrete benefit (``20\% off your next order''), paired with strong calls to action. It better suits the audience profile and earns top scores for both audience fit and marketing strength.

% INDEX_2514
\begin{figure}[h!]
  \centering
  \begin{tcolorbox}[title=\textbf{GPT-4o Response by \texttt{EvaluateAgent} for Active Old Followers}, colback=gray!5, colframe=gray!40!black, fontupper=\small]

    Audience Match Score A: 1 \\
    Audience Match Reason A: The message references a `first order,' which does not align with the `crm\_Active\_Old\_Followers' segment, as these are existing followers likely already familiar or customers, making the offer irrelevant or confusing.\\

    Marketing Score A: 3 \\
    Marketing Reason A: The call to action is unclear and potentially misleading since it references a `first order' for a segment that likely already made orders or interacted previously. Lack of a specific discount or clear benefit reduces persuasiveness and motivation.\\

    Audience Match Score B: 5\\
    Audience Match Reason B: The message suits active old followers by acknowledging their ongoing support and offering a meaningful, clearly stated discount for their next order, relevant to established customers.\\

    Marketing Score B: 5\\
    Marketing Reason B: The message is clear, concise, and motivating with a strong call to action and a specific 20\% discount, creating urgency with `limited time,' making it persuasive and effective.
    
  \end{tcolorbox}
  \caption{Illustrative example of \texttt{EvaluateAgent} response (audience match and marketing scores) for the active old customer audience segment.}
  \label{fig:output_evaluate_agent_old_1}
\end{figure}

For the potential new customer segment, Figure~\ref{fig:output_evaluate_agent_new_2} shows that \texttt{EvaluateAgent} strongly prefers Message B. This message avoids assuming any prior customer relationship and instead focuses on driving action through urgency (``limited-time steals'') and clarity (``guaranteed savings''). These are precisely the types of persuasive techniques that appeal to new customers who may be unfamiliar with the brand and require strong incentives to convert. In contrast, Message A's tone---``we really missed you''---is clearly targeted at returning users and thus mismatched for this audience. The agent correctly identifies this discrepancy and selects the more appropriate and effective version.

% INDEX_51_prefer
\begin{figure}[h!]
  \centering
  \begin{tcolorbox}[title=\textbf{GPT-4o Response by \texttt{EvaluateAgent} for Potential New Customers}, colback=gray!5, colframe=gray!40!black, fontupper=\small]

    Preferred Message: B \\
    Reason: Message B is more persuasive and better suited for the `crm\_potential\_new\_customers' segment because it creates a sense of urgency with phrases like `Limited-Time Steals' and `Hurry and grab these limited-time deals before they're gone.' It emphasizes scarcity and immediate action, which are effective in motivating new customers to make a purchase. Additionally, the message is concise, clear, and highlights guaranteed savings, which appeals strongly to potential new customers looking for value. In contrast, Message A feels more like a re-engagement message for existing customers (`It's been a while since we last saw you'), which is less relevant for potentials who may not have interacted with the shop before.
    
  \end{tcolorbox}
  \caption{Illustrative example of \texttt{EvaluateAgent} preference decision for the potential new customer audience segment.}
  \label{fig:output_evaluate_agent_new_2}
\end{figure}

In the case of active old followers (Figure~\ref{fig:output_evaluate_agent_old_2}), \texttt{EvaluateAgent} again prefers Message B for its audience fit and marketing clarity. Message B directly acknowledges the customer's continued support and provides a well-specified incentive---``20\% off your next order''---that feels both personal and actionable. Meanwhile, Message A refers vaguely to a ``first order'', which can confuse or alienate users who are already loyal. The agent's decision highlights the importance of aligning promotional language with customer lifecycle stage---what works for new users may backfire with returning ones.

% INDEX_2514_prefer
\begin{figure}[h!]
  \centering
  \begin{tcolorbox}[title=\textbf{GPT-4o Response by \texttt{EvaluateAgent} for Active Old Followers}, colback=gray!5, colframe=gray!40!black, fontupper=\small]

    Preferred Message: B \\
    Reason: Message B is more persuasive and better suited for the `crm\_Active\_Old\_Followers' segment because it emphasizes exclusivity and urgency with phrases like `Exclusive Deal Just for You' and `Limited Time,' which can motivate immediate action. It also offers a clear, concrete discount (20\% off your next order), which is more compelling than the vague `special offer for your first order' in Message A. Since the audience consists of active old followers, highlighting a discount on the next order (rather than a first order) acknowledges their prior engagement, making the message feel more personalized and relevant.
    
  \end{tcolorbox}
  \caption{Illustrative example of \texttt{EvaluateAgent} preference decision for the active old customer audience segment.}
  \label{fig:output_evaluate_agent_old_2}
\end{figure}

%%%%%%%%%%%%%%%%%%%%%%%%%%%%%%
\section{Template Quality Analyses} 

\begin{figure}[h!]
    \centering
    \includegraphics[width=1\linewidth]{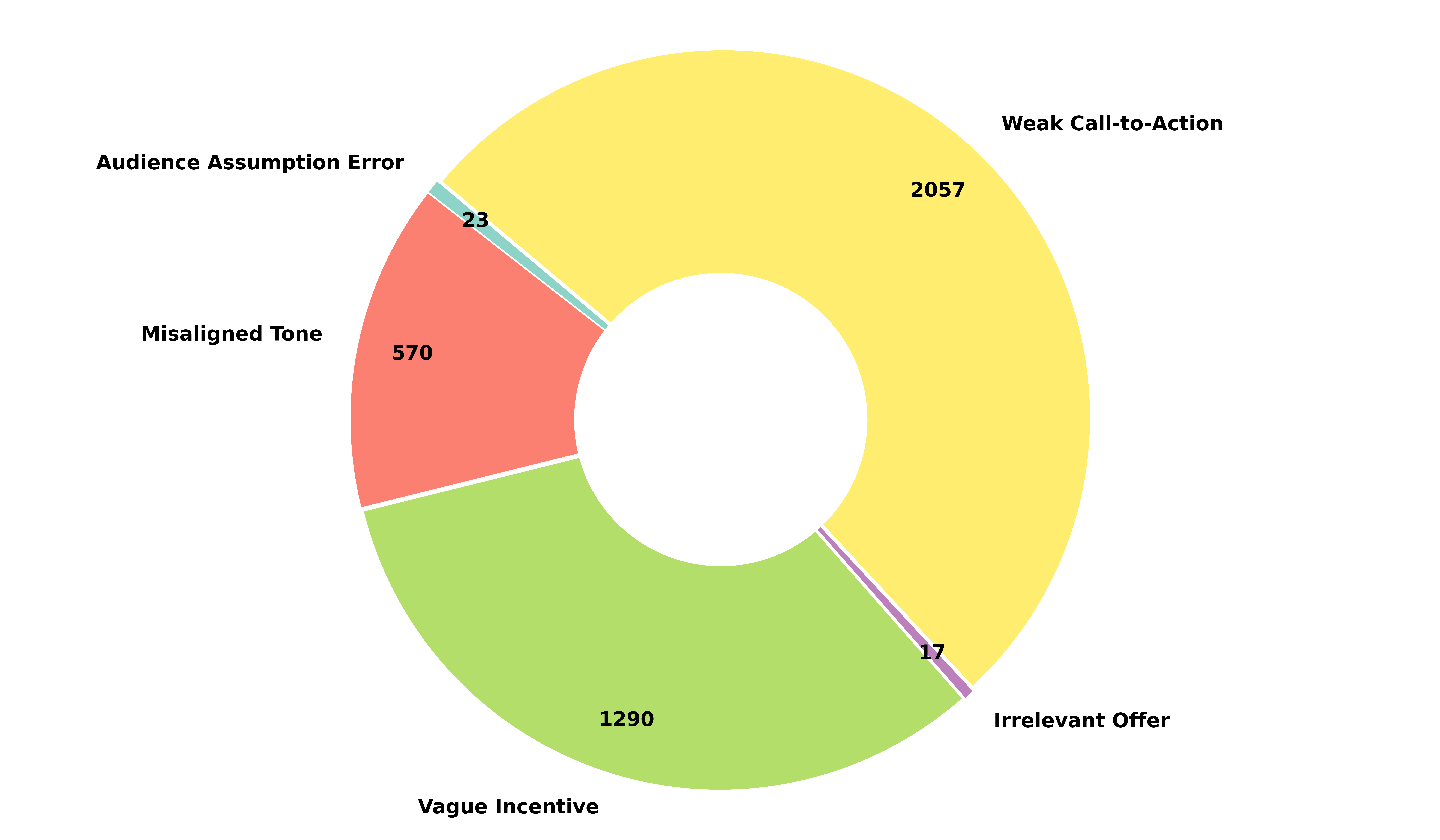}
    \caption{Distribution of quality issues in original CRM templates based on automated evaluation using o3.}
    \label{fig:quality_donut}
\end{figure}

We conduct a systematic analysis of CRM message template quality using our custom evaluation sheet (Appendix~\ref{appendix:evaluation_sheet}). Each template is assessed by o3 \citep{openai2025o4mini} and assigned exactly one error type according to predefined criteria. Figure~\ref{fig:quality_donut} presents the distribution of identified errors.

We find that the most common issue is Weak Call-to-Action (2,057 cases), where messages fail to provide a clear or urgent prompt for user action. Vague Incentive appears frequently as well (1,290 cases), often reflecting a lack of concrete or detailed value propositions. Misaligned Tone accounts for 570 cases, indicating mismatches in emotional or stylistic alignment between message and audience. Audience Assumption Errors and Irrelevant Offers occur less frequently but still signal problematic assumptions about customer history or behavior. This motivates the development of \texttt{TemplateAgent}, which is guided by these insights to generate templates that better match audience needs and communication goals.

% %%%%%%%%%%%%%%%%%%%%%%%%%%%%%%
\section{Instruction Sheet for Template Review}
\label{appendix:evaluation_sheet}

\textbf{Provided Materials}
\begin{itemize}
    \item Original Template: Each example includes only the original poor performance CRM message template.
    \item Audience Segment: The target customer segment for the message is specified (e.g., potential new customers, active old followers).
    \item Voucher: If applicable, the corresponding voucher information is provided (e.g., voucher title, discount type).
    \item Product: If applicable, the corresponding promoted product information is included (e.g., product name or category).
\end{itemize}

\textbf{Steps to Follow}

\begin{itemize}
    \item Read the message carefully.
    \item Categorize the problem using the error types below.
    \item Provide short feedback explaining your choice.
\end{itemize}

\textbf{Error Types and Criteria}

\begin{itemize}
    \item Audience Assumption Error
    \begin{itemize}
        \item Description: Assumes incorrect prior engagement with the user.
        \item Indicators: References to past interaction for new customers; offers for first-time purchase to repeat customers.
        \item Example: ``Thanks for following!'' sent to non-followers.
    \end{itemize}

    \item Weak Call-to-Action
    \begin{itemize}
        \item Description: Lacks clarity or urgency in motivating the user to act.
        \item Indicators: Generic phrases; no time-sensitive or action-oriented language.
        \item Example: ``Shop now'' with no clear reason or benefit.
    \end{itemize}

    \item Vague Incentive
    \begin{itemize}
        \item Description: Mentions a promotion without specifying what it is.
        \item Indicators: Use of terms like ``special offer'' with no details on discount or benefit.
        \item Example: ``Here's a deal for you!'' without content.
    \end{itemize}

    \item Misaligned Tone
    \begin{itemize}
        \item Description: The tone does not match the audience's familiarity level.
        \item Indicators: Overly casual or formal; assumes emotional connection without basis.
        \item Example: ``We miss you!'' to someone who never interacted.
    \end{itemize}

    \item Irrelevant Offer
    \begin{itemize}
        \item Description: The offer does not fit the customer's likely status or behavior.
        \item Indicators: Offering first-order discounts to frequent buyers.
        \item Example: ``Your first order!'' for loyal customers.
    \end{itemize}

\end{itemize}

\textbf{Example Application}

\begin{itemize}
    \item Audience Segment: Potential New Customer
    \begin{itemize}
        \item Message: ``It's been a while---check out our bestsellers!''
        \item Identified Error(s): Audience Assumption Error
        \item Feedback: This implies a prior relationship which does not exist. Rewrite to welcome the new visitor and highlight value.
    \end{itemize}

    \item Audience Segment: Active Old Follower
    \begin{itemize}
        \item Message: ``Thanks for following---enjoy your first-order discount!''
        \item Identified Error(s): Irrelevant Offer, Vague Incentive
        \item Feedback: This audience is not new; avoid ‘first order' wording. Clearly state what the benefit is.
    \end{itemize}
    
\end{itemize}

\end{document}